# Recognition of handwritten MNIST digits on low-memory 2 Kb RAM Arduino board using LogNNet reservoir neural network

Y A Izotov[1,3], A A Velichko[1], A A Ivshin[2] and R E Novitskiy[2]

[1] Institute of Physics and Technology, Petrozavodsk State University, 33 Lenin str., 185910, Petrozavodsk, Russia
[2] K-SkAI LLC, premises 20, 17 Naberezhnaya Varkausa, 185031, Petrozavodsk, Russia

[3] E-mail: izotov93@yandex.ru

**Abstract.** The presented compact algorithm for recognizing handwritten digits of the MNIST database, created on the LogNNet reservoir neural network, reaches the recognition accuracy of 82%. The algorithm was tested on a low-memory Arduino board with 2 Kb static RAM low-power microcontroller. The dependences of the accuracy and time of image recognition on the number of neurons in the reservoir have been investigated. The memory allocation demonstrates that the algorithm stores all the necessary information in RAM without using additional data storage, and operates with original images without preliminary processing. The simple structure of the algorithm, with appropriate training, can be adapted for wide practical application, for example, for creating mobile biosensors for early diagnosis of adverse events in medicine. The study results are important for the implementation of artificial intelligence on peripheral constrained IoT devices and for edge computing.

## 1. Introduction
Neural networks occupy a leading position in the processing of large amounts of information in various fields of science and technology, for example, in recognition of medical images [1], modelling [2] and control [3]. The particular importance of the analysis of data sets in real time using artificial intelligence technologies can be illustrated by the early diagnosis of adverse events in medicine [4], where the information about the patient's condition is transmitted from biosensors to receiving devices using the concept of "Internet of Things" (IoT) [5] and data transmission networks between physical objects. The rapid growth of the global market for the Internet of Things of Medical Devices (IoMT) [6], improved confidentiality, high speed of real time data processing, significant reduction of technology cost, and improved quality of outpatient medical services drive a global paradigm shift in healthcare. Decentralization of medical care and a shift from "cloud" technologies for processing medical data to "Edge computing" and "Fog computing" [7] are expected in the near future. The integration of IoT and artificial intelligence (AI) will allow for intelligent pre-processing of data on the peripheral devices without sending information to the cloud, reflecting the key area of the contemporary scientific research in the field of the Internet of things [8] and edge computing [9]. The significant research task is the development of new neural network technologies that could be successfully implemented in the IoT [10], widely used in daily activities, in healthcare [11], in designing robots [12] and in a smart homes [13], while consuming less computing resources and RAM memory.

The research on deep learning networks based on complex schemes using the convolutional filters is actively progressing [14]. Such neural networks require significant computing resources not only during training, but also during operation. An alternative for deep learning neural networks are neural networks based on reservoir computations [15]. The main idea of the reservoir computing is to use the recurrent neural network as a reservoir with rich dynamics and powerful computing capabilities. In this case, the reservoir is formed randomly, and it eliminates the need to conduct its training the in most cases. Only the output neural network, the classifier, is trained. The most popular networks are echo state networks (ESN [16]) and liquid state machines (LSM [17]). Until recently, reservoir networks did not find practical implementation on devices with low computational resources and memory, as storing a significant array of weights was a major problem. To address this issue, we have proposed the architecture of the neural network LogNNet [18], which uses a recurrent logistic mapping that describes a model of deterministic chaos. The chaotic filters, which affect the incoming information, are generated during the operation of the algorithm. As a result, the information is chaotically mixed, however, the valuable signs, which are not visible from the beginning, are extracted from the data. Reservoir neural networks use a similar mechanism. With a simple feedforward network architecture, where signals propagate exclusively from input to output, LogNNet is easy to program and install on hardware platforms. Chaotic filters are created during program operation and consume an insignificant amount of RAM.

Compact algorithms are usually tested on well-known databases such as MNIST-10 (the database of handwritten numbers [19]), CIFAR10/100 (the database of photographs of various objects [20]), Chars4K (the database of graphic symbols [21]). LogNNet was tested on MNIST-10 database and demonstrated a recognition accuracy of ~ 96.3% [18], whereas the LogNNet architecture used no more than 29 Kb of RAM. Other compact algorithms that operate on peripheral devices with 2-16 Kb RAM are algorithms based on the GBDT decision tree [22], Bonsai [23], and ProtoNN [24], which uses the kNN method. It is worth to mention the algorithm that implements a convolution neural network and uses only 2 Kb RAM [25], demonstrating a classification accuracy of ~ 99.15% on MNIST-10 database. Algorithms based on the architecture of the recurrent network Spectral-RNN [26] and FastGRNN [27] achieve ~ 98% classification accuracy on MNIST-10 with a model size of ~ 6 Kb. All mentioned algorithms have a complex structure, and their adaptation for the user's tasks requires significant efforts.

Arduino boards are effectively used to automate processes in many areas, such as education [28], medicine [29], chemical production [30], and resource monitoring in the IoT [31]. A project on an Arduino board for MNIST-10 recognition is described in the literature [32], which uses a linear classifier with two hidden layers and achieves the recognition accuracy of ~ 96%. This result is comparable to the results for the LogNNet network 784:100:60:10 [18]. However, the method in project [32] applies preliminary down sampling of images, and it requires third-party computing resources in addition to the Arduino board.

In the current study, we investigate the capabilities of the LogNNet neural network algorithm for recognizing handwritten digits from the MNIST-10 database on an Arduino UNO board with an extremely low ~ 2 Kb RAM (in datasheet SRAM). The images are received in the original form, and the image recognition happens entirely on the board. The algorithm is simple to implement and can be used for scientific and educational purposes. The research results have the application potential for the tasks of integrating IoT and AI.

## 2. Architecture of the neural network LogNNet

The LogNNet architecture developed in [18] is demonstrated in Figure 1. Its configuration is designed to recognize handwritten digits with a size of 28 × 28 pixels from the MNIST-10 database (available on Yan LeCun's Internet page [19]). An image of a digit is fed into the network input, and then the image is transformed into a one-dimensional array $Y$ using the T-pattern transformation. We used the simplest template shown in Figure 1 with per-column image input. An array $Y$ is fed to the input of the reservoir. The array contains the values of the pixel intensities normalized to 1, has 784 values in total, and the zero offset element is $Y[0] = 1$.

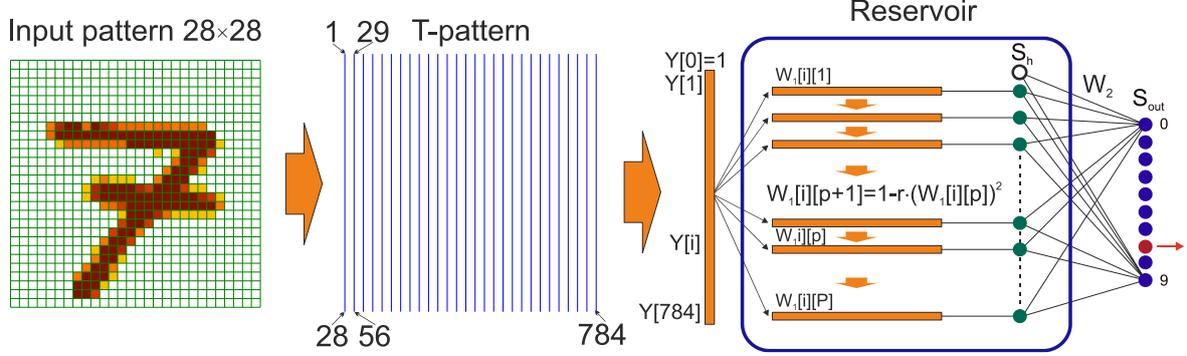

**Figure 1**. LogNNet neural network architecture.

The reservoir and the output neural network operate as feedforward networks, according to the equations

$$S_h = f_h(Y \cdot W_1), \qquad (1)$$

$$S_{out} = f_{out}(S_h \cdot W_2). \qquad (2)$$

In equations (1) and (2), $S_h$ are the neurons of the hidden layer located in the reservoir, $S_{out}$ are the neurons of the output layer, $Y$ is the array of input data, $W_1$, $W_2$ are the matrix of weight coefficients, from the input to the hidden layer and from the hidden layer to the output layer, respectively, $f_h$ is the identical activation function, with subsequent normalization of values in the range from -0.5 to 0.5, and $f_{out}$ is the logistic activation function.

The matrix $W_1$ is set using the equations (3-4):

$$W_1[i][1] = A \cdot \sin\left(\frac{i}{784} \cdot \frac{\pi}{B}\right), \qquad (3)$$

$$W_1[i][p+1] = 1 - r \cdot W_1[i][p] \cdot W_1[i][p], \qquad (4)$$

where $r$, $A$ and $B$ are constants, $i$ and $p$ are variables, $i = 0\text{-}784$, $p = 1..P$, $P$ is the number of neurons in the hidden layer.

Equation (4) is a logistic mapping where the next value is calculated based on the previous one. The equation is commonly used in population biology, and is an example of a simple equation for calculating a sequence of chaotic values. Equations (3-4) can fill the matrix $W_1$ of any dimension with chaotic numbers. The elements of the $W_1$ matrix are sequentially calculated during the network operation, and occupy a small amount of RAM memory. This approach saves device memory; however, it additionally loads the processor with calculations and increases the network operation time.

The matrix of coefficients $W_2$ is trained using the backpropagation method on the MNIST-10 database. The database contains 60,000 training handwritten images and 10,000 test images of numbers from 0 to 9, in the form of a matrix of 28 × 28 pixels. The brightness of a pixel is represented by a value from 0 to 255.

The normalization coefficients of the function $f_h$ are calculated from the values of $S_h$ using the first 1000 elements of the training set. The maximum, minimum, and weighted mean values of $S_h$ for each neuron of the hidden layer were used as normalization coefficients.

The constants *r, A* and *B* were selected by optimization methods using the particle swarm method, where the fitness function was the function of the neural network recognition accuracy. The values of these constants are given in the appendix (Figure A2).

After determining all the parameters, the trained neural network was programmed on the Arduino UNO board.

## 3. Implementation of the LogNNet network on the Arduino UNO board

The objective of the study is to test the LogNNet architecture for recognizing handwritten digits from the MNIST-10 database on a microcontroller with an extremely low amount of RAM. For the implementation of the LogNNet neural network and its subsequent testing, a low-memory 2 Kb RAM board Arduino UNO with an ATmega 328P microcontroller is chosen. Table 1 shows the main characteristics of the board.

**Table 1. Characteristics of the Arduino UNO board.**

| Characteristic | Value |
| --- | --- |
| SRAM | 2048 bytes |
| Flash memory | 32 Kbytes |
| Microprocessor | ATmega 328P |
| Clock frequency | 16 MHz |
| Weight | 46 g |
| Board dimensions | 69×53×20 (mm) |

In [18], we demonstrated that with an increase in the number of neurons in the hidden layer of the *P* reservoir, the recognition accuracy of MNIST-10 digits increases. The main limiting factor for the number of neurons in the hidden layer, when implementing LogNNet on the ATmega 328P microcontroller, is the amount of static RAM. The mathematical calculations for the network operation require a large number of variables of various types that take up RAM memory. The optimal configuration of the neural network LogNNet 784:20:10 was selected, where the numbers indicate the parameters of the network: $S = 784$ - the size of the input data array *Y*, $P = 20$ - the number of neurons in the hidden layer $S_h$, and $M = 10$ - the number of neurons in the output layer $S_{out}$ that determine the result of the network operation.

A fragment of the executable program code describing the main "void loop" is presented in Figure 2a, and the functions and procedures used are shown in Figure A1 in Appendix. The program is written in the C programming language in the Arduino IDE development environment, and has a compact form in the editor, occupying 62 lines and 5934 bytes of the device's flash memory. All necessary constants (*S, P, M*), values of the weights of the $W_2$ arrays and normalization parameters are contained in the library file LogNNet.h (see Figure A2, Appendix) and are loaded into the RAM memory at the start of the program. At the start of the program, the serial port of the board is initialized by the procedure Serial.begin(115200) in the lines 45-47 of the "void setup" code block, as demonstrated in Figure 2a.

The testing process is performed by transferring images to the board one by one through the serial port. The transferred images have the original scale of 28 × 28 pixels, and are not pre-processed.

The algorithm of the program's operation checks in a loop the presence of data in the input buffer via the Serial.available function and reads all 784 bytes of the *Y* array. Next, the neural network response is calculated.

At the first stage, a preliminary calculation of the neurons of the hidden layer takes place, the code for the Hidden_Layer_Calculation procedure is shown in Figure A1 (a) (see Appendix, lines 7-19). The procedure calculates the weight coefficients $W_1$ using Equaitons (3, 4), then the arrays *Y* and $W_1$ are multiplied in accordance with Equation (1). The values of the hidden layer neurons are averaged and normalized for the final calculation. The normalization procedure is called Hidden_Layer_Normalization and its code is demonstrated in Figure A1 (a) (see Appendix, lines 21-27). The procedure uses the minimum (*minH*), maximum (*maxH*) and weighted average (*meanH*) parameters from the LogNNet.h library file. These values are pre-determined during network training, and they are floating point numbers. The values are written to the microcontroller memory in integer form (to save RAM), and they are divided by a scale factor for the reverse conversion to the "float" type. For arrays *minH* and *maxH*, the scale factor is 100, and for the *meanH* array, the coefficient is 10000. The final values of the neurons of the hidden layer are stored in the $S_h$ array. The length of the $S_h$ array is 21, and it contains 20 hidden layer neurons calculated in the Hidden_Layer_Normalization procedure and 1 bias neuron.

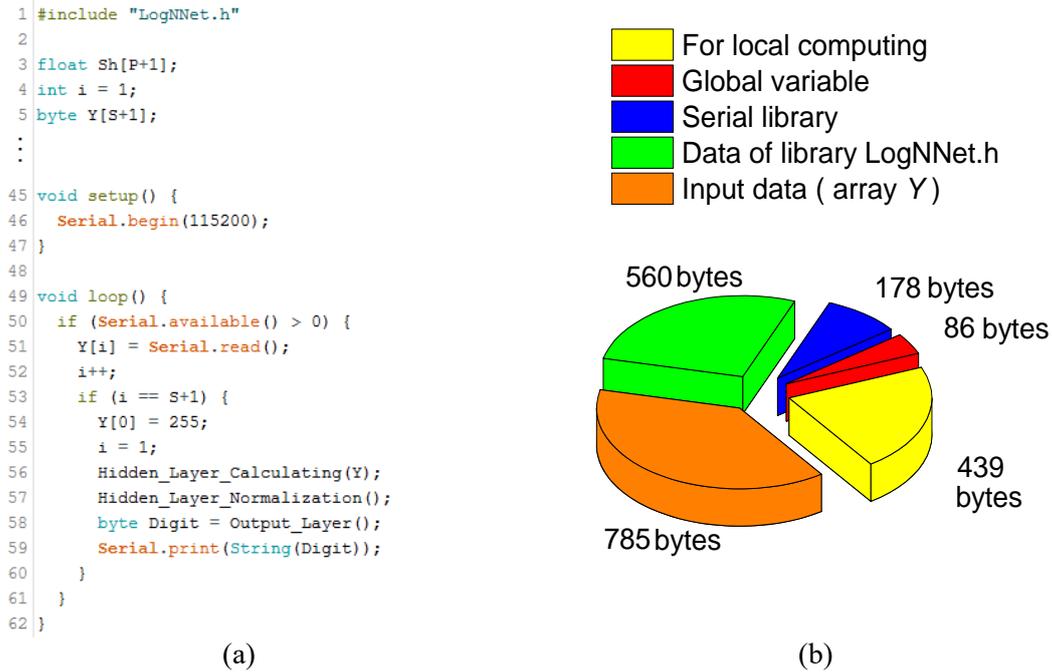

(a)            (b)

**Figure 2**. A fragment of the executable program for the Arduino UNO (a), and the allocation of RAM memory of the microcontroller board for the configuration LogNNet 784:20:10 (b).

The neurons of the output layer are calculated in the Output_Layer function, and the result of the neural network operation is obtained; the code is demonstrated in Figure A1 (b) (see Appendix, lines 29-43). The function calculates the values of the $S_{out}$ array and outputs the most probable digit corresponding to the recognized image, in the range from 0 to 9 (variable *Digit*). The function uses an array of weighting coefficients $W_2$, which was obtained during preliminary network training on a computer. The $W_2$ array is stored in the controller RAM in integer form and is loaded from the LogNNet.h library. The scale factor of the $W_2$ array equals to 100.

*3.1. Memory allocation.*
The allocation of the microcontroller's RAM is presented in Figure 2b. Global variables $S_h$ and $i$ occupy 86 bytes in RAM (4.2%). The data of the LogNNet.h library are loaded at the start of the program and occupy 560 bytes (27.34%). The program uses a library for working with a serial port (Serial), which occupies 178 bytes (8.68%). Input data from the serial port is written to array *Y*, which is 785 bytes (38.33%). As a result, with 20 hidden layer neurons, the volume of permanently used RAM is 1609 bytes, and the microcontroller has 439 bytes left for mathematical calculations.

*3.2. Neural network training and testing.*
    The training of the neural network LogNNet is performed on a computer.

The process of network testing consists of sequentially downloading each of the 10,000 pictures of the MNIST-10 database test set through the serial port, calculating $S_{out}$ and receiving the *Digit* result back to the computer. This technique allows full test of the network on the entire test set, without loading the set into the board's memory.

### 4. Results and discussion

Table 2 presents the results of the MNIST-10 image recognition accuracy using the LogNNet network with a different number of neurons *P* in the reservoir, and the processing time of one image by the Arduino UNO board. The image recognition accuracy on a personal computer and on the Arduino UNO board differ after the 2nd decimal place of the result number. It may be explained by the method of storing the $W_2$ weights and normalization parameters in integer form, when significant digits of numbers are lost. In addition, Arduino UNO uses the float type numbers with 8 significant digits for calculations, while the computer software (Delphi programming language) operates with the real type numbers with 16 significant digits. Nevertheless, such good accuracy is a significant result and allows preliminary testing of the network on a computer with more processing power.

The highest recognition accuracy corresponds to the network configuration with the number of neurons in the reservoir *P* = 20. The accuracy reaches values of ~ 82% and is comparable, for example, with the NeuralNet Pruning network, which has an accuracy of ~ 81% [33].

An increase in the number of neurons in the reservoir (*P* > 20) does not give a noticeable increase in the classification accuracy and requires additional expansion of RAM memory. For each additional neuron, 30 bytes of RAM are added to increase the arrays $S_h$, *minH, maxH, meanH*, and $W_2$. A further accuracy increase can be achieved by internal preprocessing of the array *Y* or by increasing the number of hidden layers of the output classifier. These ideas can be as a subject for future research.

**Table 2. Results of LogNNet operation with different number of neurons *P* in the reservoir.**

| Configuration of network LogNNet *S*:*P*:*M* | Recognition accuracy of MNIST-10 image on a computer | Recognition accuracy of MNIST-10 image on Arduino UNO board | Recognition time of one image on Arduino UNO board |
|---|---|---|---|
| LogNNet 784:20:10 | 82.04 % | 82.03 % | 7.11 s |
| LogNNet 784:18:10 | 80.46 % | 80.45 % | 6.01 s |
| LogNNet 784:15:10 | 76.89 % | 76.94 % | 4.06 s |
| LogNNet 784:10:10 | 70.00 % | 70.01 % | 2.06 s |

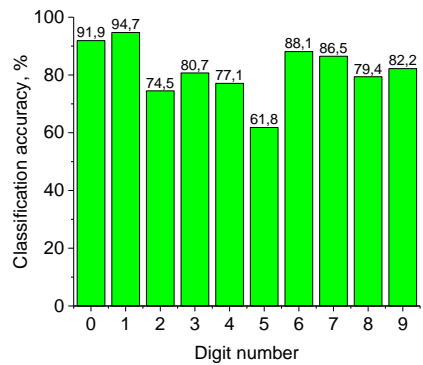

**Figure 4**. Distribution of classification accuracy by digits for LogNNet 784:20:10.

The distribution of the classification accuracy of individual digits from the MNIST-10 base for LogNNet 784:20:10 is shown in Figure 4. The neural network recognizes digits with different probability. For the digit "1", the probability is the highest and equals to ~ 94.7%, or, in other words, 769 images out of 1032 possible images are correctly recognized. The network showed the worst result when recognizing the digit "5", when the probability was ~ 61.8% out of 892 images. The similar reduced classification accuracy of the digit "5" was observed in other studies, for example, in Hopfield network [34]. The reason for this may be the similarity of the digit "5" to digits "3", "6" and "9".

The recognition time for one image on the board is several seconds. The time is mainly determined by the duration of the sequential calculation of the $W_1$ array elements in the Hidden_Layer_Calculation procedure.

The recognition time can be reduced by applying the faster Algorithm 2 described in [18], but algorithm implementation requires a microcontroller with a large amount of RAM memory. Such processing speed of input data is acceptable for private practical applications that do not require a quick response, for example, in environmental monitoring systems or medical monitoring. The range of tasks to be solved by LogNNet is not limited to recognizing the MNIST database images, and can be expanded to other areas after corresponding retraining.

## 5. Conclusion

The study presents a compact algorithm for recognizing handwritten digits of the MNIST-10 database, created on the basis of the LogNNet reservoir neural network. The algorithm is tested on the low-memory Arduino UNO board, which has a low-power microcontroller and 2048 bytes of RAM. Classification accuracy of ~ 82% was achieved with a small ($P = 20$) number of neurons in the reservoir. The program code is compact, simple in structure and can be used in the scientific and educational processes. With appropriate training, the presented LogNNet algorithm can be adapted for wide practical applications, for example, for creating mobile biosensors for early diagnosis of adverse events in IoMT, and implemented on a wide range of hardware and software, for example, Raspberry PI, Orange Pi, Arduino, Microduino and Femtoduino. The results facilitate the creation of prototypes of peripheral devices with low RAM, performing required functions using own artificial intelligence in the space of the Internet of Things.

**Acknowledgment**

## Appendix A

```
6
7  void Hidden_Layer_Calculating(byte Y[S+1]) {
8    float W1 = 0;
9    Sh[0] = 1;
10   for (int j = 1; j <= P; j++) {
11     Sh[j] = 0;
12     for (int i = 0; i <= S; i++) {
13       W1 = A * sin(i/float(S)*PI/B);
14       for (int k = 2; k <= j; k++)
15         W1 = 1 - (r * W1 * W1);
16       Sh[j] = Sh[j] + Y[i]/255.0 * W1;
17     }
18   }
19 }
20
21 void Hidden_Layer_Normalization() {
22   Sh[0] = 1;
23   for (int j = 1; j <= P; j++)
24     Sh[j] = ((Sh[j] - minH[j-1]/100.0) /
25       (maxH[j-1]/100.0 - minH[j-1]/100.0))
26       - 0.5 - meanH[j-1]/10000.0;
27 }
```

```
28
29 byte Output_Layer() {
30   float Sout[M];
31   for (int j = 0; j < M; j++) {
32     Sout[j] = 0;
33     for (int i = 0; i <= P; i++)
34       Sout[j] = Sout[j]+Sh[i]*W2[i][j]/100.0;
35     Sout[j] = 1 / (1 + exp(-1*Sout[j]));
36   }
37   byte digit = 0;
38   for (int i = 0; i < M; i++) {
39     if (Sout[i] > Sout[digit])
40       digit = i;
41   }
42   return digit;
43 }
44
```

(a)  (b)

**Figure A1**. The procedures Hidden_Layer_Calculation, Hidden_Layer_Normalization (a) and the Output_Layer function (b) used for calculating the LogNNet network on the Arduino UNO.

```
1  #pragma once
2  #include <Arduino.h>
3
4  #define S 784
5  #define P 20
6  #define M 10
7
8  #define A 0.672304710111946
9  #define B 1.57740808024101
10 #define r 1.86376633343066
11
12 int maxH[P] = {16458, 13985, 16194, 4846, 9755, 12284, 5592, 5826, 9223,
13                6653, 3504, 11273, 3993, 5204, 7919, 3889, 9294, 6141, 5063, 5334};
14 int minH[P] = {1043, 944, 900, -2416, -1539, -2423, -2509, -816, -761, -2896,
15                -2205, -125, -2282, -457, -17, -2211, -112, -276, -858, 37};
16 int meanH[P] = {-2188, -2301, -2094, -595, -1707, -1204, -667, -1419, -1395,
17                 -846, -1025, -1562, -451, -1382, -1471, -1016, -1929, -1348, -1137, -1274};
18 int W2[P+1][M] = {
19                  {-491, -915, -453, -353, -455, -419, -516, -584, -324, -431},
20                  {211, -522, 120, -463, -509, -115, 96, -412, 344, 16},
21                  {17, -1044, 775, -133, -1325, -149, -105, -806, -539, -741},
22                  {304, 56, -933, -890, 535, -152, 283, 229, 1362, 965},
23                  {-95, -38, -2354, -4, 965, 219, -1500, 1769, 1154, 1590},
24                  {-620, 4, -1564, -710, 2701, -1338, 1485, -657, 331, 897},
25                  {-1012, 1187, 1633, -1741, -1798, -1191, 126, 667, 2425, 46},
26                  {-644, -242, 391, -1942, -334, -64, 745, -131, 914, 838},
27                  {-317, -523, -594, -825, -988, 991, 1085, -257, 373, 99},
28                  {105, 565, -1483, 1147, 614, 1539, -364, -508, -389, -1703},
29                  {-819, 331, 196, 2209, -900, -340, -471, 118, -781, -684},
30                  {1250, -2406, -340, 925, -580, 416, -98, -850, -727, -664},
31                  {-200, -1120, -480, 1527, -516, 245, -685, 68, -1278, -331},
32                  {137, 542, -203, 464, -111, 323, -479, -399, -237, -77},
33                  {7, -352, 312, -839, 778, 836, -501, -154, 1087, 220},
34                  {712, 1207, 1556, -872, 431, 501, -752, -1718, 668, -1003},
35                  {602, -996, 745, -123, 965, -1148, -1411, 1678, -923, -888},
36                  {1558, -856, 14, 410, -1222, -1103, 752, -14, -1345, 602},
37                  {-305, -712, 645, 913, -46, -1199, -969, -469, 86, 727},
38                  {-588, -72, 655, 733, -20, -930, -197, 501, -1088, -244},
39                  {-541, -1298, -427, 564, 599, 342, 218, 997, -947, -444}
40 };
```

**Figure A2.** LogNNet.h library source code.

The source code version is available in the GitHub repository:
**https://github.com/izotov93/LogNNet_Arduino**